\documentclass[10pt,twocolumn,letterpaper]{article}

\usepackage{cvpr}
\cvprfinalcopy
\usepackage{amssymb}
\usepackage{amsmath}
\usepackage{graphicx}
\usepackage{graphics}
\usepackage{subcaption}
\usepackage{multirow}
\graphicspath{ {images/} }
\begin{document}

%%%%%%%%% TITLE
\title{A DNN Framework For Text Image Rectification From Planar Transformations}

\author{
Chengzhe Yan, Jie Hu, Changshui Zhang\vspace{4mm}\\
% \and
{Tsinghua University}\\
}

\maketitle
\begin{abstract}
In this paper, a novel neural network architecture is proposed attempting to rectify text images with mild assumptions. A new dataset of text images is collected to verify our model and open to public. We explored the capability of deep neural network in learning geometric transformation and found the model could segment the text image without explicit supervised segmentation information. Experiments show the architecture proposed can restore planar transformations with wonderful robustness and effectiveness.
\end{abstract}

\section{Introduction}
\label{sec:intro}
In computer vision and machine learning community, text image rectification\cite{liang2008geometric,ye2015text} from planar transformation have long been a topic of concern. Often a system of OCR and image information retrieval has a pipeline with several steps, including rectification, detection and recognition\cite{ye2015text} helping improve performance.

\begin{figure}[htp]
  \centering
  \includegraphics[width=0.46\textwidth]{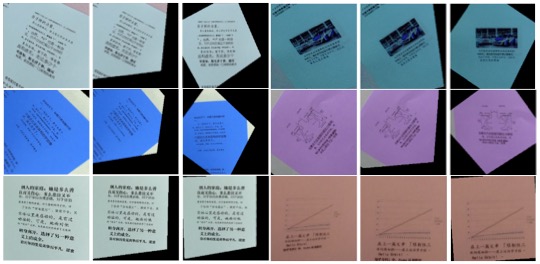}
  \caption{Sample rectifications performed grouped in 3 columns: first column is distorted samples, middle shows rectification on perspective, and rotation rectification are presented in last columns.}
  \label{fig:samplecorr}
\end{figure}

Unlike other object recognition problem, text line patches in text images usually have large aspect ratio as shown in Figure~\ref{fig:samplecorr}. This property becomes problematic when image is distorted with planar transformation like rotation and perspective. A horizontal bounding box based algorithm usually pick out the lines with low percentage of interest pixels from distorted images.

Some solutions attempt to solve recognition with distortion by data augmentation with more labeled samples. It's very expensive to collect and label large dataset with different rotational and perspective distortions. Besides, more complicated models are required to learn on larger dataset, which bring more utilization of computation and storage resources in training and application. Synthesis data are recently proposed and used in learning scene text images\cite{jaderberg2014synthetic}. For Chinese characters, this would be difficult and impractical to overcome this problem using data augmentation due to the scale of charset.

An orthogonal way is to learn its transforming parameters and recover its original view as processed by human. Methods worked in this way usually have assumptions like some borderline exists, camera parameters available\cite{cambra2011towards} or low rank property of text images\cite{zhang2012tilt} which may not be satisfied in many situations when noise and complex textures exist.

Since the emergence of deep neural network(DNN)\cite{krizhevsky2012imagenet,lecun1998gradient,lecun2015deep}, it gradually reshapes the methodology of machine learning by pushing the state of the art of different tasks much forward, and bring out novel applications\cite{simonyan2014very,ren2015faster}. Convolution neural network\cite{lecun1995convolutional} with deep architecture has proven to be top performance in many computer vision tasks\cite{girshick2015fast,Zhang_2016_CVPR}. In text images analysis, deep methods for OCR using and LSTM\cite{graves2006connectionist} has shown superior performance in recognition but may fail with planar transformation.

Recently, some researches try to solve affine and perspective transformation using deep neural network. Among them, one research is spatial transformer network(STN)\cite{jaderberg2015spatial} which attempts to improve classification accuracy by inserting a transformer layer into an end-to-end neural network. It could detect and transform features learned into better view hence increase accuracy. However, one of its drawbacks is its inability to rectify images from planar transformation stably as human beings.

In this paper, a new framework of DNN is proposed combining the advantages of STN and supervised learning to learn rectification of planar transformations as shown in Figure \ref{fig:samplecorr}. The only assumption is that existence of a few parallel human readable text lines in the context of rectified images. We give transformation parameters directly to DNN, training it to learn rectification parameters in stage-wise method. Classification of rotation degrees is used instead of regression by discretizing range of rotation into intervals\cite{sun2015learning}. Initialization of convolution kernel is different with commonly used methods\cite{glorot2010understanding} to achieve better performance.

The advantages of this model include no only much milder assumption with no formats or prior knowledge of datasets. Deep models have much better robustness, and adaptive to different variations than traditional image processing methods. Besides, we found although no segmentation labels is directly provided to the model, it learns to distinguish part of the input image into different types according to context. It seems the model have mastered learning to focus on meaningful elements and regions with rectification training.

This model is benchmarked on a new dataset collected containing Chinese texts of different types and illumination conditions. We use real world data collected and transformed with generate parameters for training and testing. The outcome indicates the effectiveness and robustness of the proposed architecture.

\section{Methods and Models}
\label{sec:arch}
\subsection{Background, Notation and Formulation}
Planar transformations are common in image processing and can be modeled as a perspective transformation. Define $I$ as original image, $I'$ as transformed image. Each pixel $p$ in $I$ with coordinate $(x,y)$ in one image is rectified and interpolated to $p'$ with coordinate $(x',y')$ in $I'$. Here define related parameters including rotation as $\theta$, scaling factor $\alpha$, perspective parameters $(P_x, P_y)$ and translation of $(T_x,T_y)$. Planar transformation can be formulated as:
\begin{equation}
\label{eq:perspective}
  \begin{bmatrix}
    \alpha\cos(\theta) & \alpha\sin(\theta) & T_x\\
    -\alpha\sin(\theta) & \alpha\cos(\theta) & T_y\\
    P_x & P_y & 1
  \end{bmatrix}
\begin{bmatrix}
  x\\
  y\\
  1
\end{bmatrix}
=\begin{bmatrix}
  \hat{x}\\
  \hat{y}\\
  \hat{z}
\end{bmatrix},\\
\end{equation}

\begin{equation}
x' = \hat{x}/\hat{z},y' = \hat{y}/\hat{z}.
\end{equation}

Most recognition algorithms use bounding boxes and object detection methods to locate text area which is able to deal with different $(T_x,T_y)$ and $\alpha$. While some algorithms can handle problem of non-horizontal text lines, most of them require parallel text lines\cite{koo2012text}, or simply separately locate each word or neighboring texts\cite{Zhang_2016_CVPR}. With perspective transformation, parallel condition will disappear and a crossing point of these lines can be estimated using image processing method with strong assumptions as surveyed in \cite{ye2015text}. 

\subsection{Parameter Entangling and Stage-wise Model}
Given the label data, it would be straight forward to build a CNN and solve it as a regression problem. The capability of feature learning, high representative ability and robustness to noise of CNN make it best choice for this task. However, it's hard to estimate all the parameters simultaneously due to the nonlinearity of the mapping to approximate. Rotation parameters are serious interfered by perspective transformation as parallel property disappears as shown in (\ref{eq:perspective}).

Also, parameters of transformation are in multiple orders of magnitude. This problem is defined as ill-condition problem\cite{boyd2004convex} and is hard for neural networks to learn\cite{saarinen1993ill}. Since most existing deep learning research focus on tasks with discrete labels, they mainly concern numerically about initialization and gradient vanishing for nonlinear activation functions\cite{glorot2010understanding}.

Though difficult to regress parameters simultaneously, we could build models in stage-wise style inspired by past pipeline based research. If we eliminate perspective distortion first, and recover the parallel property of text lines, estimation of $\theta$ would be feasible. For a mathematic expression, we can decompose (\ref{eq:perspective}) into:
\begin{equation}
  \begin{bmatrix}
    1 & 0 & 0\\
    0 & 1 & 0\\
    P_x & P_y & 1
  \end{bmatrix}
  \begin{bmatrix}
    \alpha\cos(\theta) & \alpha\sin(\theta) & T_x\\
    -\alpha\sin(\theta) & \alpha\cos(\theta) & T_y\\
    0 & 0 & 1
  \end{bmatrix}
  \begin{bmatrix}
    x\\
    y\\
    1
  \end{bmatrix}
  =\begin{bmatrix}
    \hat{x}\\
    \hat{y}\\
    \hat{z}
  \end{bmatrix}.
\end{equation}

Assuming the position text area not biased much over the image, right-down corner element varies quite a little from 1. It means, transformation is equivalent to translation first, then rotation and perspective transformation in order. We rectify it in the inverse order, by learning $P_x$ and $P_y$ first, then estimating rotation angle $\theta$, and locating the text area at last. There have been lots of research done on locating the text areas, here we mainly solve the first 2 steps.

Transformation parameters in these 2 steps are learned using 2-stage training model. The CNN architecture proposed in this paper rectify perspective and rotation transformation parameters for images with text in 2-stages using loss function  
\begin{equation}
  \label{eq:loss}
  \min \sum_{i=1}^n||P_i-\mathcal{P}(X_i)||_2^2-\lambda\sum_{i=1}^n\sum_{j=1}^Ct_{ij}\log(\mathcal{A}(X_i)_j).
\end{equation}
In (\ref{eq:loss}), $X_i$ is the $i$th sample in the dataset. $P_i$ is a tuple $(P_{xi},P_{yi})$ as the corresponding perspective parameter. $t_{ij}$ is one-hot expression of angle parameter in predefined discretized precision which in our paper set to 2 degrees. The loss function consists 2 parts: the first part is $\mathcal{P}$, a neural network aiming at learning to estimate $P_x$ and $P_y$. While the second part is an angle classifier $\mathcal{A}$ to estimate $\theta$ using cross-entropy loss instead of regression. 

Supervised STN is used to connect these 2 components and form an end-to-end neural system, which will be introduced in section \ref{subsection:sstn}. We give explicit rectification value to DNN and use its prediction as input of STN, hoping this end-to-end model could better understand the geometric concept from the perspective of human.

\subsection{Perspective Learning}
\label{subsec:persp}
\begin{figure}[htp]
  \centering
  \includegraphics[width=0.4\textwidth]{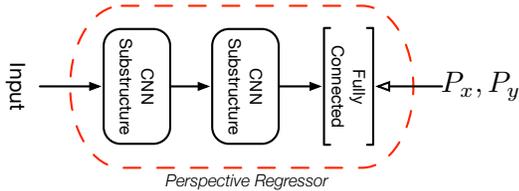}
  \caption{Architecture of DNN learning $P_x$ and $P_y$.}
  \label{fig:archpsp}
\end{figure}
In traditional methods, researchers studied the relationship between different indicators and $P_x$ and $P_y$. There indicators either summarize local features, or attempt to find the vanishing point brought by the perspective transformation, which measures level of parallel distortion. In estimation of $P_x$ and $P_y$, they are on the same magnitude, hence its Jacobian matrix is not ill-conditioned any more. Therefore, we can numerically approximate the mapping using fully connected neural networks.

Input sample images pass through a few convolution substructure formed by convolution, pooling and batched normalization layer. Features extracted from CNN are used by fully connected layer to approximate $P_x$ and $P_y$.

\subsection{Rotation Learning}
\label{sec:angle}
\begin{figure}[htp]
  \centering
  \includegraphics[width=0.4\textwidth]{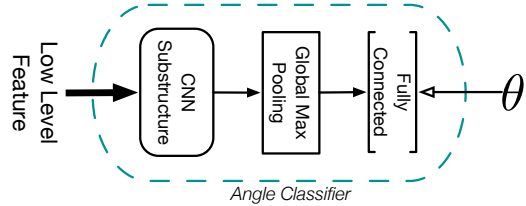}
  \caption{Architecture of DNN learning $\theta$.}
  \label{fig:archrotation}
\end{figure}
To get accurate prediction, it's intuitive to regress $\theta$ on low level features as a continuous variable. $\ell_2$ norm, which is defined as $||x||_2=\sqrt{\sum |x_i|^2}$ is usually used to learn the regression. However, after several layers of pooling, straight lines become zigzag shape in digital images. And difference between close samples are so small on feature map in value of losses. Therefore this mapping has high nonlinear property and hard to regress use classical neural network, especially difficult to identify the proper structure of hidden layer.

It is a general method to discretize continuous variable into disjoint intervals. That is using intervals labels as the surrogate of continuous value. Despite that we use discrete labels as output, what we care is not the accuracy, but the regression residue error since we are estimate an ordinal not cardinal value.

Clearly, discretize the range into infinite number of intervals is equivalent to continuous regression. Actually we can improve effectiveness of classifier as an estimator of $\theta$ by adding other penalty term. To prove this, a penalty term aiming at minimize the distance between ground truth and prediction is added, and use $\mu$ to tradeoff between $\ell_2$ loss and cross entropy. Classification accuracy of this additional loss term has no significant difference with (\ref{eq:loss}), but variance of errors decreases. Cross entropy makes no assumption on prior distribution while the $\ell_2$ term partly introduce Gaussian prior into estimation. From another view, this term add penalty to large bias from ground truth.

\subsection{Supervised Spatial Transformer Network}
\label{subsection:sstn}
\begin{figure}[htp]
  \centering
  \begin{subfigure}[htp]{0.35\textwidth}
    \includegraphics[width=\textwidth]{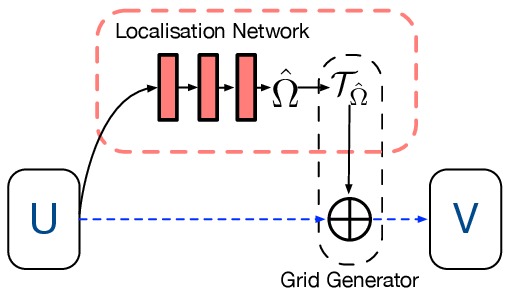}
    \caption{STN}
    \label{fig:stn}
  \end{subfigure}
  \begin{subfigure}[htp]{0.37\textwidth}
    \includegraphics[width=\textwidth]{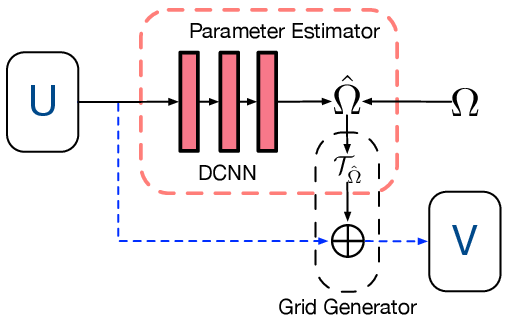}
    \caption{Supervised STN}
    \label{fig:sstn}
  \end{subfigure}
  \caption{STN and Supervised STN. $\Omega$ is the transformation parameter, while $\hat{\Omega}$ is an estimator of $\Omega$.}
\end{figure}
Angle classification in our model need features free from perspective transformation. An end-to-end neural network is more appropriate and usually have better performance. We proposed supervised STN to connect stages to an integral system.

In overcoming the problem of transformation, researchers introduce spatial transformer network\cite{jaderberg2015spatial} to DNN by inserting one or several transformer layer into the stacked layers as shown in Figure~\ref{fig:stn}. The transformer layer is a multilayer neural network whose input is feature map from previous layers' output, while its output is a set of parameters which could describe the transformation it learned. No other information is fed into the layer, and features used for learning only comes from back-propagation gradient of classification loss. After epochs of training, the transformer are capable of transforming feature maps best for classification.

However, its transformation is different from human vision since no information on concept of geometric rectified from human perspective is provided for training. In practice, we find the performance of STN is quite sensitive to the ratio of object area respective to the image size. If the object of interest is small relative to image size, it fails to locate the object, nor estimation of other transformation parameters.

The difference between transformation of STN and human vision is still an obstacle for better recognition and understanding of document images. Supervised STN proposed in this paper shown in Figure~\ref{fig:sstn}, connect the hidden layer of localization network of STN with labels of transformation parameter $\Omega$, and put it as a component of final loss function. This method enables the neural network to obtain an estimator $\hat{\Omega}$ learning rectifications from humans' perspective. In other words, supervised information is provided to the hidden layer of STN, enabling these layers to learn its specified objective. For even larger neural network with more objectives, it can be used as a essential part aiming at transform the features into more proper spaces.

Like STN, several supervised STN can be inserted into network layers with different loss components. If the transformation is not feasible with single stage, we can decompose the transformation into several stages, and arrange each stage with corresponding subset of label values. In this way, transformation with high complexity can be approximated and eliminated, helping to build an end-to-end neural system.

\subsection{Integral Architecture}
\begin{figure}[htp]
  \centering
  \includegraphics[width=0.48\textwidth]{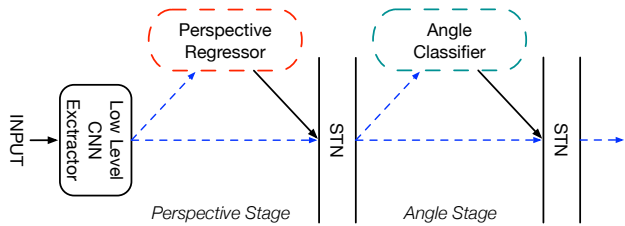}
  \caption{Architecture of DNN learning transformation parameters in 2 stages.}
  \label{fig:arch}
\end{figure}
The primary architecture of model we used in this paper is shown in Figure~\ref{fig:arch}. Each stage is a supervised STN using the respective transformation parameters as labels and targets. $P_x$ and $P_y$ are estimated in the first stage, using CNN describe in section \ref{subsec:persp}. The low level feature map produced by convolution is transformed by the supervised STN, and feed into the second stage. In back-propagation, if the parameters in first stage is trained with acceptable accuracy, the feature map can be inversely transformed backward to provide gradient to transformation parameters in the first stage.

Inside the network, dropout\cite{srivastava2014dropout} is applied to increase generalization. A batch normalization\cite{ioffe2015batch} layer is inserted after each pooling layer and connected it with next convolution substructure with positive effect. STN receives transformation parameters from regression result of $P_x$ and $P_y$ and input feature map output by front layers, then inversely transforms the input tensor with linear interpolation.

On estimation of $\theta$, output feature map from 1st stage STN are processed with convolutions and pooling for the first. The input image went through STN of both stages in succession with $P_x,P_y$ and $\theta$ given as output to produce the rectified image.

This model, unlike traditional separated pipeline framework, put each component together to form an end-to-end DNN. Training such an integral system altogether would benefit from enhancement of combined feature learning. 

\subsection{Convolution Kernel Setting}
% \label{subsec:convolution}
% \begin{figure}[!htp]
%   \centering
%   \includegraphics[width=0.4\textwidth]{kernel}
%   \caption{kernel of different size with initialization proposed.}
%   \label{fig:kernel}
% \end{figure}

Angle detection is a special task. Unlike most tasks that have much diversified samples, like classification between birds and chair, rotation of the same object of arbitrary shape will have nearly the same output feature map. But for strings of text in image, we can take it as a flat rectangles or wide lines. To reduce number of parameters in model, global average layer\cite{szegedy2015going} like GoogleNet are chosen instead of dense layer. This structure helps improving generalization ability with much less model parameters.

If the convolution kernel is a tilted banded matrix which has direction $\theta$, with nonzero elements in banded and others setting to zero. Then convolve it with a tilted banded matrix, then its convolution will be maximized on that location if they are tilted in the same direction. If they perfectly match, then the maximum element in the convolution output should be the maximum among all candidate angles. This initialization could help the system find a better starting point, and the mathematic derivation make it not a surprise that global average pooling using maximum method have better performance than fully connected model. This reduces much of model parameters and increase its generalization ability.

Another important hyper-parameter is the kernel size of angle convolution. Kernel of smaller size have less representative capability of direction. Drawbacks of larger $k$ include worse generalization as more parameters are involved in model. Besides, the time complexity of convolution which is $O(c'cmnk^2)$ increasing in quadratic order of $k$ and scale of model parameters also increase quadratically. We need to balance between less computation, better generalization, and representative power.

% Given a feature map of $X$, estimation of $\theta$ can be formulated as
% \begin{equation}
% \label{eq:anglec}
%   \hat{\theta}=\arg\max_{\theta\in \Omega}\{\max(X\circ \mathcal{K}(\theta))\}.
% \end{equation}

% In (\ref{eq:anglec}), $\mathcal{K}$ is mapping from angle $\theta$ to a tilted banded matrix whose normal vector is rotated $\theta$, and $\mathcal{A}$ is the true mapping from image to angle.

\section{Experiments}
\subsection{Dataset}
\label{sec:data}
As far as we know, no public text image rectification dataset is appropriate for this task. We need the dataset collected follows several properties: 1). No marks, borders or boundaries of regular shape in patches exists unless they occur in the context of image content; 2). Text comes from large scale charset with more diversified patterns; 3) Lines of text should vary in font, fontsize, and other settings. Diversification on illumination condition and camera parameters is also necessary.

The dataset are divided into captions and texts samples. The caption samples contain a image and a paragraph of text under it, and the font and font-size of text are the same for each single patch. For text images, we generate random setting for each very line, and draw them on each patch.

Six patches are put in a $2\times3$ grid and drawn on a large sample image. Four different kind of marks are drawn on the four very corner of samples images leaving enough space to keep generated patches free from marks, so that patches used for training contain no interfering marks. 

We then printed these samples on different type of paper, including plain A4 papers of different colors, and kraft paper, card paper with different texture types. These printed samples are paragraphed using cameras with different configuration under various illumination conditions.

These original digital images are manually labeled on marks around the corners and aligned. 243 images, including 158 texts and 85 of captions are collected. Using preserved location values, we get 6 patches from each sample. For each patch, we generate several random transformation parameters of translation, scaling, rotation and perspective transformations and applied on it. These generated parameters and sample patches works as target values and samples for training and verification. Some sample patches are shown in Figure~\ref{fig:samplecorr}. The only difference of this synthetic dataset with real world data comes from interpolation used when applying the transformations, which is not sensitive to deep models.

\subsection{Implementation Details}
\label{sec:impl}
\begin{table}[htp]
  \caption{Architecture details}
  \label{tab:detailA}
  \centering
  \begin{tabular}{c|l|c|l}
  \hline
  \textbf{A} & \textbf{1st stage} & \textbf{B} & \textbf{2nd stage} \\
  \hline
     1 & conv(2,64,3,3) & 1& conv(128,128,3,3)\\
     2 & pool(2,2) & 2 & pool(2,2)\\
     3 & batch-norm & 3 & conv(n,128,k,k)\\
     4 & conv(2,128,3,3) & 4 & global\\
     5 & pool(2,2) & 5 & fc(512)\\
     6 & batch-norm & 6 & softmax(n)\\\cline{3-4}
     7 & conv(1,128,3,3) & 1 & conv(2,64,3,3)\\
     8 & pool(2, 2) & 2 & pool(2,2)\\
     9 & batch-norm  & 3 & conv(2,128,3,3)\\
     10 & conv(1,128,3,3) & 4 & pool(2,2)\\
     11 & pool(2, 2)& 5 & conv(128,128,3,3)\\
     12 & batch-norm & 6 & pool(2,2) \\
     13 & fc(3,1024) & 7 & conv(n,128,k,k)\\
     14 & fc(1, 2) & 8 & global \\
     15 & STN(15, 8) & 9  & fc(512) \\
      &  & 10 & softmax(n) \\
  \hline
  \end{tabular}
\end{table}

We establish DNN architecture as shown in Tab.~\ref{tab:detailA}, where $n$ stands for number of angle intervals, conv$(m,n,t,t)$ stands for m connected convolution layers of $n$ channel with kernel size $t$, STN(A, B) means a STN layer with A as transformation parameters and B the feature map to transform, and $k$ the chosen kernel size. We will use indexes in Tab.~\ref{tab:detailA} to indicate layers in later description. 

Our input image size is 256 pixels wide. Fully connected layers act as approximation learner, and we need to balance the scale of them. Also it is necessary to guarantee that the output feature map is not too small for angle convolution layer detecting angles. We choose 4 convolution-pooling layer at last to keep model parameters scale appropriate.

Larger kernel size could improve accuracy of angle estimation, but decrease generalization power and increase time for training and validation. We finally choose 9 for B3. We choose $\lambda$ as 1.0, to balance between 2 component of loss.

We train regression NN firstly and use trained model parameters for angle classifier. Choice of learning rate is more art than science that neither large or small is appropriate: larger learning rate will make learning oscillate far from optimal while with small ones training may stuck on bad local minimums\cite{glorot2010understanding}. We finetune learning rate carefully from $\{1e^{-4}, 2e^{-5}, 1e^{-5}\}$.

% Adam self-adaptively adjusts learning rate for each parameter, in practice we keep learning rate constant for tens of epochs, and observe the performance on training set and validation set. Starting from default configuration of Adam, once it shows overfitted on validation set, we change to smaller learning rate. When training angle classifier, we treat each stage differently by applying different learning rate for each part of parameters because front convolution layers has been pretrained in perspective stage.

\subsection{Performance and Analysis}
\label{sec:perf}
The experiment is running on a server with GTX 1080 GPGPU and Xeon-2630v3 CPU. For this paper, we use 10000 patches in the dataset and break into 8000 for training, 1000 for validation, and 1000 for test. We generate $P_x$ and $P_y$ in uniform distribution with interval $[-9e^{-4},9e^{-4}]$, which give largest parallel distortion of 24 degrees. $P_x$ and $P_y$ are transformed using $1e^4\times(P+1e^{-3})$, as ReLU can only give positive response, and inversely transformed in STN. Angles for rotation of different scale are generated and used to test whether its learning ability is correlated with entangling transformation parameter range. $T_x,T_y$ and $\alpha$ are generated with guarantee that the generated patch will not cross the boundaries after all four transformation are applied.

We conduct many experiments with different model configurations and hyper-parameter settings. Comparison between these settings could give more detailed understanding about the method and its functioning principles.

\subsubsection{Perspective Regression}
Some researches for estimation of $P_x$ and $P_y$ need assumptions on range of $\theta$, or find better performance in small scale range. Estimation of $P_x$ and $P_y$ is the fundamental part of entire system, hence we need to check whether the range of $\theta$ have large effect on estimation of $P_x$ and $P_y$.

We generate dataset with different scale of $\theta$, in $\{\pi/3,\pi/2,2\pi/3\}$. If learning capability is limited by scale of rotation, their performance should be quite different. We give $\ell_2$ loss and $\ell_1$ bias of regression result, which is shown in Tab.~\ref{tab:perspective}.
\begin{table}[h]
  \caption{Experiment Result on $P_x$ and $P_y$ Regression}
  \label{tab:perspective}
  \centering
  \begin{tabular}{l|cc|cc}
  \hline
  \textbf{Expr.} & \multicolumn{2}{c|}{\textbf{$\ell_2$}} & \multicolumn{2}{c}{\textbf{$\ell_1$}} \\
  \hline
         & train & val & train & val\\
     $\pi/3$ & 0.7873 & 1.9140  & 0.0848 & 0.2217 \\
     $\pi/2$ & 0.8163 & 1.9370 & 0.0887 & 0.2156 \\
     $2\pi/3$ & 0.8407 & 1.8729 & 0.0872 & 0.2152 \\
  \hline
  \end{tabular}
\end{table}
Figures in Tab.~\ref{tab:perspective} imply the generalization capability under different angle scales. The outcome varies yet within a regular range. With these results, we can conclude that deep regression have nice generality in different $\theta$ scale. But there are still bias on validation dataset, whether it's accurate enough to estimate $\theta$ remains to be tested.

\subsubsection{Angle Classification}
\begin{table*}[htp]
  \caption{Expr. Angle Classification}
  \label{tab:aclass}
  \centering
  \begin{tabular}{|l|cc|cc|cc|cc|}
  \hline
  \multirow{2}{*}{\textbf{Expr.}} & \multicolumn{2}{c|}{\textbf{Acc}} & \multicolumn{2}{c|}{\textbf{Var}} & \multicolumn{2}{c|}{\textbf{Top~2}} & \multicolumn{2}{c|}{\textbf{Top~5}} \\\cline{2-9}
     & Valid & Test & Valid & Test & Valid & Test & Valid & Test \\\hline
    Shared & 49.60\% & 47.00\% & 0.6340 & 0.6860 & 80.0\% & 78.6\% & \textbf{98.7}\% & 99.0\\
    Independent & 48.40\% & 43.50\% & 0.9650 & 0.9250 & 77.9\% & 78.0\% & 97.3\% & 97.8\%\\
    $\ell_2$ Reg. & \textbf{53.50}\% & \textbf{47.10}\% & \textbf{0.5920} & \textbf{0.6690} & \textbf{83.0}\% & \textbf{79.9}\% &  98.5\% & \textbf{99.1}\%\\
  \hline
  \end{tabular}
\end{table*}
As introduced in section \ref{sec:impl}, training of angle classifier are based on model parameters trained in the perspective stage. We designed experiments on verify angle classifier in different settings. Experiment result on angle classification are shown in Tab.~\ref{tab:aclass}. We choose parameters with best benchmark performance according to its results on training and validation dataset, and test its capability on test dataset. Models in comparison include original shared kernel model, independent kernel model, and loss with $\ell_2$ penalty. The consistence and effectiveness of $\theta$ estimation are analyzed. Furthermore, we investigate more on output by analyzing top-k accuracy, which means accuracy of closest prediction within the k largest output of softmax.
\paragraph{Shared versus Independent Kernels}
\begin{figure}[tb]
  \centering
  \includegraphics[width=.42\textwidth]{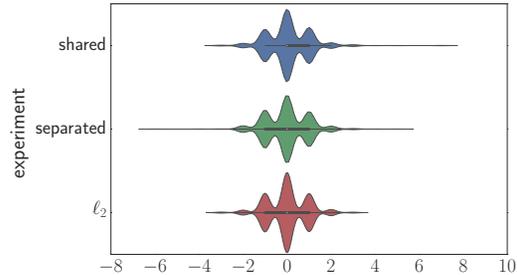}
  \caption{Violin plot of angle classification result. The band draws smoothed distribution, and inside the band is the boxplot of corresponding indicator.}
  \label{fig:violin}
\end{figure}
As we explained in previous section, kernel parameter sharing reuse convolution kernels in front layers. Besides reducing model parameters, the collaborated training process of these two tasks helps the model to learn better features for estimation of $P_x,P_y$ and $\theta$. Compare the first 2 rows in Tab.~\ref{tab:aclass}, we found parameter sharing bring improvement in faster convergence and better accuracy: in same epochs of training, it get higher training accuracy compared with independent model and its performance on validation and test dataset also show its better generalization power. 

Also, as shown in Tab.~\ref{tab:aclass}, variance of prediction error, altogether prove that shared kernel model have better effectiveness. We demonstrate the property of error in violin plot as Figure~\ref{fig:violin}. Judging from the distribution, prediction error is more concentrated to 0 in shared kernel model. And Tab.~\ref{tab:aclass} also indicated its better robustness in all entries.

The nature of this result could be attributed to difficulty of training large neural network for angle prediction. Using shared kernels will have optimization of front layers to start from a better initial point. But for independent model, gradient propagated to front layers would have little effect since differences between samples of neighboring angles is quite small after pooling. It's then difficult to learn appropriate features for this task.

\paragraph{Classification With Additional $\ell_2$ Term}
Another experiment focus on the impact of $\ell_2$ term. Based on the analysis, we expect adding an $\ell_2$ term in the loss function will give larger penalty to large bias. From another view, the $\ell_2$ term assume a Gaussian prior. This constraint, if meaningful, would help improve the consistence and effectiveness of $\theta$ estimation.

As shown in Tab.\ref{tab:aclass}, within same period of training, although result on training dataset implies model without $\ell_2$ have better performance, on validation and test dataset, model with $\ell_2$ penalty have achieved best result. However, their outcomes are pretty close. The difference on test dataset are within 3\% at most. Indicators in Tab.~\ref{tab:aclass} also show similar conclusion. 

This comparison evaluation is reasonable, since the prior knowledge and penalty maybe not helpful in learning such a complicated mapping. Choice of $\mu$ doesn't affect much since loss component of cross-entropy and estimation of $P_x$ and $P_y$ only takes $1/5$ if we choose $\mu=0.01$.

\subsubsection{Integral Performance}
\paragraph{Internal Evolution Mechanism}
\begin{figure*}[htp]
  \centering
  \includegraphics[width=0.78\textwidth]{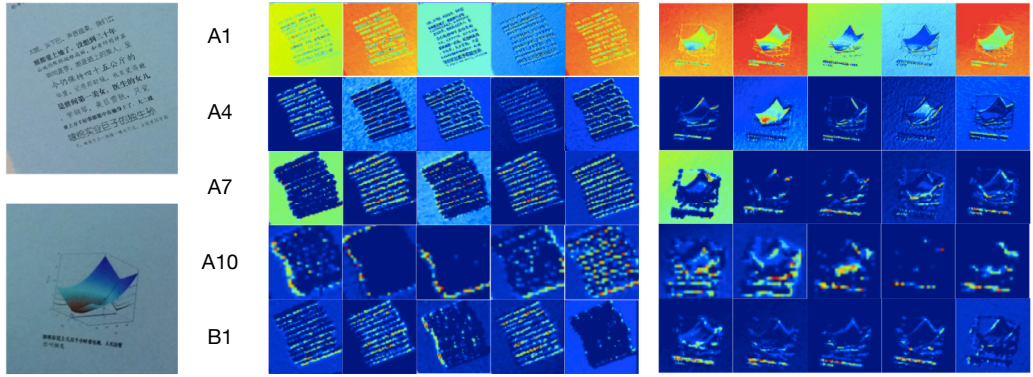}
  \caption{Feature map output by convolution layer.}
  \label{fig:featuremap}
\end{figure*} 
\label{subsection:inner}
There has always critical statements on deep learning about its black box property with puzzling internal mechanism. As many research using classical image processing method to rectify each transformation have been proposed, it's better to figure out how it evolves to learn rectification and how we can reference them to understand DNN's methodology.

Internal mechanism is explored by observing and comparing feature map output of middle layers. It's amazing to find even no explicit segmentation or context information is given to the DNN, it evolve itself focusing on meaningful areas: some kernels segmented input features into background, text lines, space between lines, and other meaningful non-text regions. 

Figure~\ref{fig:featuremap} visually explains this statement which contains several manually picked but typical component of output feature map. The first 4 rows come from first stage, and last row from the second. These 2 patches are representative in dataset: one contains only text lines, while the other is caption image with lines of description under the image. From front to end layer, kernels gradually learn to focus on different image elements. For 1-3 rows, feature maps describe image in different scopes, from local to larger scale, from edge detection towards a fuzzy shape contour segmentation. From the third row of Figure~\ref{fig:featuremap} of text line case, we observe that convolution tries to make a distinction between regions of text, line spacing and background. To the caption case, with complicated background to analyze, it achieves comparable result.

The way it learns to estimate $P_x$ and $P_y$ in rear layers is hard to analyze. Here some components of feature map indicate how it works in text line case: given the contour and segmentation from front layers, it locates the very upper, bottom and left most border, these line should be perpendicular or parallel lines. The bias from its proper outlook is used by the fully connection layers to estimate $P_x$ and $P_y$. In contrast, however, it's more puzzling to analyze mechanism for caption case.

For angle estimation stage, after processed by first STN, convolution in layer 2 in shared kernel seems to segment text area more significantly. In both cases, text area are different in value with other part of feature map. This implies that even with no explicit segmentation label, after learning the concept of rectification, the DNN understand roles of different elements inside images to some extent.

\paragraph{Case Analysis}
\begin{figure}[htb]
  \centering
  \includegraphics[width=0.45\textwidth]{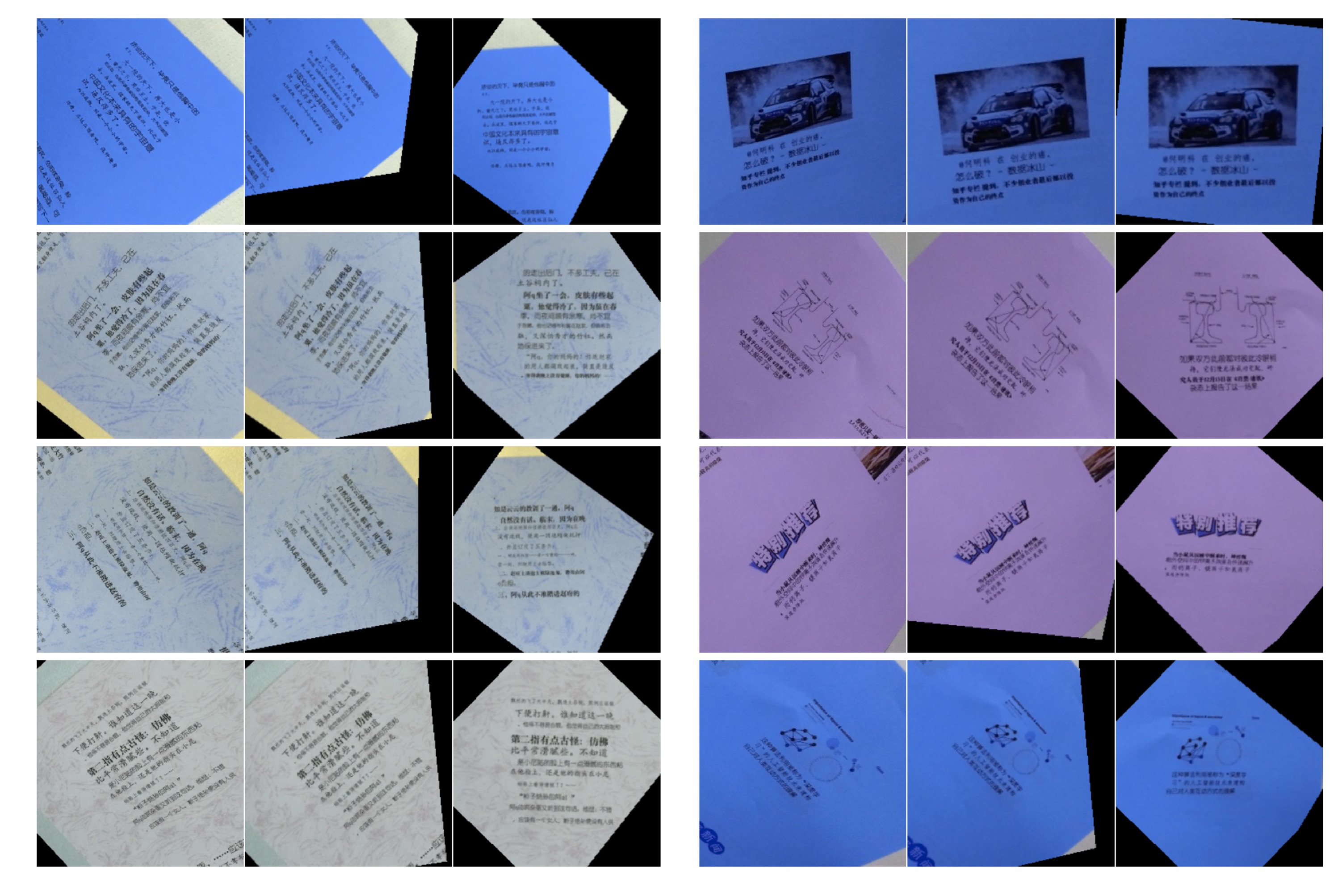}
  \caption{Some sample cases of rectification. For each group of three image from left to right is the distorted image, perspective rectified, and rotation rectified output.}
  \label{fig:good}
\end{figure}
\begin{figure}[htb]
  \centering
  \includegraphics[width=0.45\textwidth]{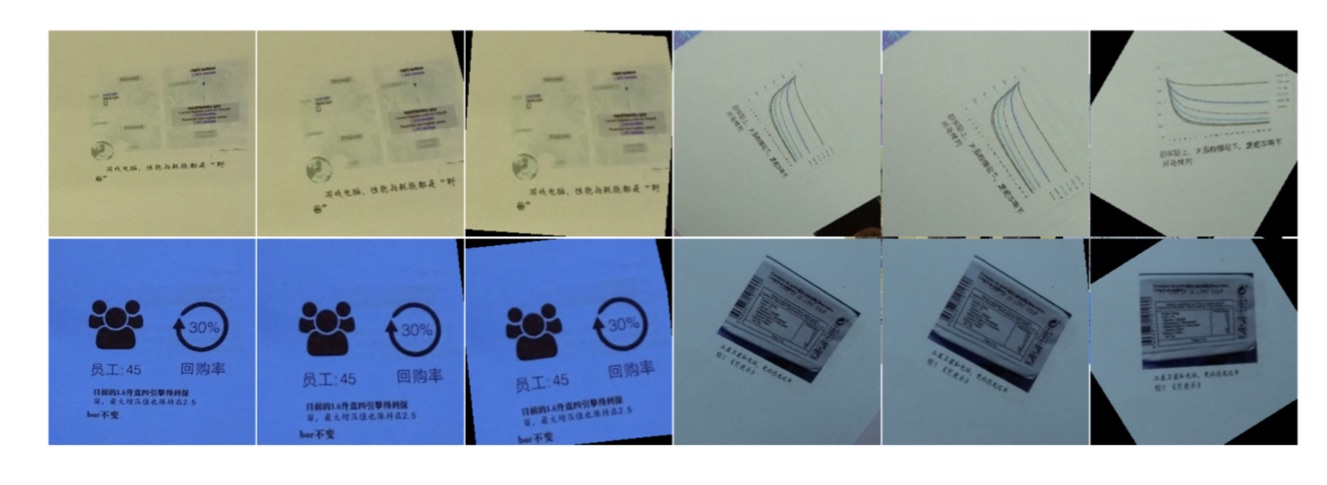}
  \caption{Problematic cases.}
  \label{fig:bad}
\end{figure}
In this part, we will give both good and bad cases in overall rectification. We choose from test datasets, and show them in order of sample, perspective rectified, and final output. Figure \ref{fig:good} shows some patches perfectly rectified including both caption and text line patches.

It matters more on bad cases plotted in Figure \ref{fig:bad}. We found our model works pretty well and stable on text line patches. But in case of caption, it sometimes fails to restore its original outlook. With detailed analysis, we guess the problem mainly exists in the regression performance in perspective stage: in caption case, with fewer lines of text, $P_x$ and $P_y$ estimation rely more on information of non-text area. Context images are different and varies in style as observed in samples, that some are geometric shapes while some others are complicated images. With such complexity, regression need more data and more various types of context images to learn more accurate and robust regression mapping.

\section{Conclusion and Future Work}
\label{sec:conclusion}
In this paper, a new deep architecture aiming at rectifying images with parallel text lines are proposed with thorough experiments comparing with different configurations and hyper-parameters. Experiments on newly collected dataset show its effectiveness and robustness.

However, even though the dataset we use have thousands of text image formats, more diversified data is needed to verify model's robustness to variations of formats. Common cases include name card, course slides, or more complicated texts in nature scenes. With more data collected, we can get better generalization capability.
\clearpage
\bibliographystyle{elsarticle-num} 
\bibliography{ref}

\end{document}